# Edge Preserving Multi-Modal Registration Based On Gradient Intensity Self-Similarity

Tamar Rott, Dorin Shriki and Tamir Bendory
Department of Electrical Engineering, Technion - Israel Institute of Technology
tamarott@gmail.com, dorinshr@gmail.com, stdory@mail.technion.ac.il

*Abstract*—Image registration is a challenging task in the world of medical imaging. Particularly, accurate edge registration plays a central role in a variety of clinical conditions. The Modality Independent Neighbourhood Descriptor (MIND) demonstrates state of the art alignment, based on the image self-similarity. However, this method appears to be less accurate regarding edge registration.

In this work, we propose a new registration method, incorporating gradient intensity and MIND self-similarity metric. Experimental results show the superiority of this method in edge registration tasks, while preserving the original MIND performance for other image features and textures.

## I. INTRODUCTION

Image registration aims to find an optimal transformation, aligning two or more images taken at different times, different points of view or by different sensors. Particularly, in medical applications it is frequently desired to register images taken in different modalities, such as computed tomography (CT), magnetic resonance imaging (MRI) and ultrasound.

Many approaches for image registration have been introduced thus far (see comprehensive survey in [14] and references therein). Mutual information (MI) has been widely applied to medical image registration [10], [15]. MI aims to find a statistical correlation across images and thereby maximizes the amount of shared information between two images. Several extensions to MI have been proposed, such as normalized mutual information [8], conditional mutual information [9] and a combination of gradient intensity and mutual information [12]. A different approach relies on structural representations of the images, such as entropy and Laplacian [16], edge detection [6], similarity-sensitive hashing [2] and morphological tools characterization [5].

The Modality Independent Neighbourhood Descriptor (MIND) proposes a multi-modal registration algorithm based on the assumption that the local structure is shared across modalities [7]. Although MIND achieves accurate and reliable alignments in a variety of registration tasks, numerical experiments show that MIND is less accurate for edge alignment, which may be substantial for several clinical conditions, including human organ recognition, tumor detection and disease diagnosis (e.g. Crohn's disease, Multiple Sclerosis). For instance, in Crohn's syndrome, accurate detection of the bowel wall is highly important for grading the severity of the disease activity [17]. Accurate estimation of tumor size and shape is crucial for determining the treatment strategy [13].

This work suggests a new image registration method, called Gradient MIND (G-MIND), incorporating gradient intensity and MIND descriptor. Leveraging the fact that the gradient emphasizes sharp brightness changes and discontinuities, our method achieves a significant improvement in edge preserving.

We show by numerical experiments on synthetic and real medical images that the G-MIND outperforms MIND in the accuracy of edge alignment, while preserving MIND performance for other features and textures.

The rest of the paper is organized as follows: Sections II and III present MIND and G-MIND algorithms respectively, Section IV is devoted to numerical experiments and ultimately Section V concludes the paper.

## II. MODALITY INDEPENDENT NEIGHBOURHOOD DESCRIPTOR

The Modality Independent Neighbourhood Descriptor (MIND) proposes a deformable registration method between two source images. The algorithm is composed of two steps. First, for each image, a descriptor is generated, relying on the self-similarity properties within small patches. Next, the similarity between the two images is defined by the difference between their descriptors, and an optimization algorithm is applied to minimize it.

The descriptor aims to find an image representation, which is modality-independent, relying on the assumption that a local structure of the underlying object is shared across modalities. In this work we focus on two-dimensional images, however, MIND can be easily extended to higher-dimensions.

Before presenting MIND, we introduce two definitions. $D_P(I,\mathbf{X}_1,\mathbf{X}_2)$ is the intensity distance between two pixels $\mathbf{X}_1$ and $\mathbf{X}_2$ of an image $I$, defined as the sum of squared differences between the intensity of all the pixels in the two patches of size $(2P+1)^2$, centered around $\mathbf{X}_1$ and $\mathbf{X}_2$. Namely,

$$D_P(I,\mathbf{X}_1,\mathbf{X}_2) = \sum_{\mathbf{p} \in \Pi} (I(\mathbf{X}_1+\mathbf{p}) - I(\mathbf{X}_2+\mathbf{p}))^2, \quad (1)$$

where
$$\Pi := [-P,\ldots,P] \times [-P,\ldots,P],$$
for some $P \in \mathbb{N}$. The variance $V(I,\mathbf{X}_c)$ is the mean of the distances between the pixel $\mathbf{X}_c$ and its four neighbours:

$$V(I,\mathbf{X}_c) = \frac{1}{4}\sum_{\mathbf{k}} D_P(I,\mathbf{X}_c,\mathbf{X}_c+\mathbf{k}). \quad (2)$$

Equipped with the definitions in Equations 1 and 2, the MIND descriptor for a two-dimensional image $I$, at a pixel $\mathbf{X}_c$ with a given neighbour $\mathbf{r}$ is defined as:

$$MIND(I,\mathbf{X}_c,\mathbf{r}) = \frac{1}{n}\exp\left(-\frac{D_P(I,\mathbf{X}_c,\mathbf{X}_c+\mathbf{r})}{V(I,\mathbf{X}_c)}\right), \mathbf{r} \in R, \quad (3)$$

where $R$ defines a search region centered at the pixel $\mathbf{X}_c$ and $n$ is a normalization constant (so that the maximum value is 1).

The complete MIND descriptor associates a self-similarity vector of size $|R|$ for each pixel $\mathbf{X}$ in the image, calculated as in Equation 3. The similarity metric between two images $I, J$ is defined as the sum of absolute differences between their corresponding MIND descriptors:

$$S(\mathbf{X}) = \frac{1}{|R|}\sum_{\mathbf{r}\in R}|MIND(I,\mathbf{X},\mathbf{r}) - MIND(J,\mathbf{X},\mathbf{r})|. \quad (4)$$

Subsequently, the Gauss-Newton optimization framework is used in order to minimize the similarity metric in Equation 4 (see Section 4.5 in [14]). By minimizing $S(\mathbf{X})$, a registration displacement $\mathbf{U}=(u_x,u_y)^T$ is obtained. The displacement can be formulated as:

$$\mathbf{X}' = \mathbf{X} + \mathbf{U}, \quad (5)$$

where $\mathbf{X}=(x, y)^T$ is transformed to the location $\mathbf{X}'=(x', y')^T$.

By finding the displacements between images $I$ and $J$, both the forward transformation of image $I$ to image $J$, and the backward transformation of image $J$ to image $I$ are computed. Ultimately, the original images are fused together into one image by averaging the forward and backward transformations.

### III. G-MIND

MIND was proven as a reliable registration algorithm for a variety of tasks (see Section 5 in [7]). However, a few registration tasks on real medical images, either taken from the Visible Human dataset [1] or provided by MIND establisher [18], were resulted in miss-alignments around edges in the image. An example is illustrated in Figure 1.

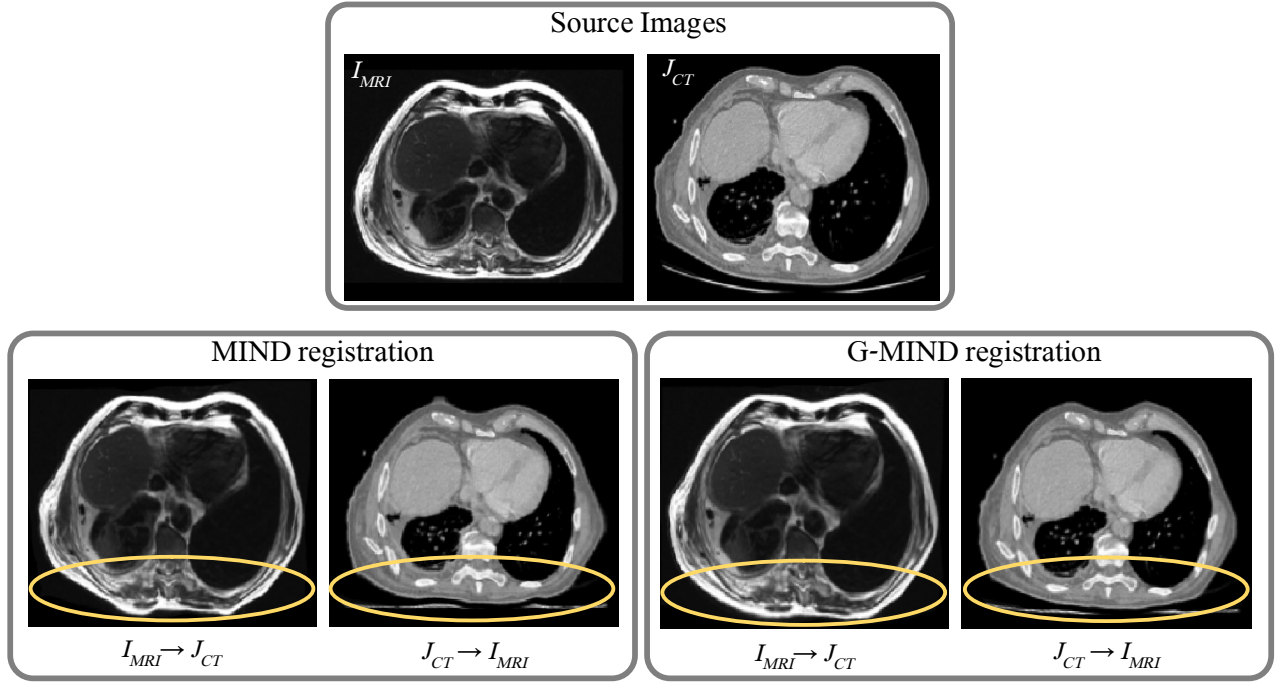

Fig. 1. Comparison between MIND and G- MIND registrations. The upper line displays MRI (left) and CT (right) source images of the upper abdomen. In the bottom line, the MIND and G-MIND registration results are shown. As can be readily seen, MIND caused a non-physiological deformation in the patient back (the bottom of the images), whereas G-MIND manages to perform a smooth registration preserving the contour of the original images.

As aforementioned, edge preserving may be crucial for several clinical conditions. We present a new registration method, called G-MIND, overcoming MIND weakness in edge registration. G-MIND incorporates the image's gradient into the MIND algorithm.

The magnitude of the gradient is determined by the sharpness of the intensity changes. Therefore, pixels with high gradient value are detected as a possible part of an edge. Additionally, the direction of the strongest change in the area surrounding the pixel is indicated by the gradient direction.

Let us denote the two gradient images $I_{\nabla \hat{x}}, I_{\nabla \hat{y}}$ of an image $I$ by:

$$\left(I_{\nabla \hat{x}}, I_{\nabla \hat{y}}\right) = \left(\frac{\partial}{\partial x}I(x,y), \frac{\partial}{\partial y}I(x,y)\right), \quad (6)$$

where $\frac{\partial}{\partial k_j}I(k_1, k_2), j=1,2$ denotes a partial derivative.

The G-MIND initially generates two gradient matrices for each of the source images $I$, $J$. Then, four MIND descriptors are generated for each of the gradient images ($I_{\nabla \hat{x}}, I_{\nabla \hat{y}}, J_{\nabla \hat{x}}, J_{\nabla \hat{y}}$) as described in Equation 3. For each direction ($\hat{x}$ and $\hat{y}$), a similarity metric is defined as follows:

$$S_{\nabla \hat{x}}(\mathbf{X}) = \frac{1}{|R|}\sum_{\mathbf{r}\in R}|MIND(I_{\nabla \hat{x}}, \mathbf{X}, \mathbf{r}) - MIND(J_{\nabla \hat{x}}, \mathbf{X}, \mathbf{r})|,$$
$$S_{\nabla \hat{y}}(\mathbf{X}) = \frac{1}{|R|}\sum_{\mathbf{r}\in R}|MIND(I_{\nabla \hat{y}}, \mathbf{X}, \mathbf{r}) - MIND(J_{\nabla \hat{y}}, \mathbf{X}, \mathbf{r})|. \quad (7)$$

Following this, Gauss-Newton optimization process is executed to minimize $S_{\nabla \hat{x}}(\mathbf{X})$ and $S_{\nabla \hat{y}}(\mathbf{X})$, resulting in the G-MIND displacements $\mathbf{U}_{\nabla \hat{x}}$ and $\mathbf{U}_{\nabla \hat{y}}$ (see Equation 5). The final displacement is reached by averaging the displacements of both directions:

$$\mathbf{U}_{G\text{-}MIND}(\mathbf{X}) = \frac{1}{2}\mathbf{U}_{\nabla \hat{x}} + \frac{1}{2}\mathbf{U}_{\nabla \hat{y}}. \quad (8)$$

The fused image is then obtained by implying the displacement $\mathbf{U}_{G\text{-}MIND}$ on the source images.

Figure 1 illustrates a comparison between MIND and G-MIND for edge registration. The upper line of the figure shows MRI and CT scans of a patient's upper abdomen. The bottom line presents the registration results of the two methods. Each one of the source images is fully transformed by the forward or backward registration transformations, as described in Section II. After the registration is performed, the MRI image should be fully spatially compatible to the source CT image and vice versa. As can be seen, MIND causes a non-physiological deformation in the back of the patient, while G-MIND performs an edge-preserving registration.

IV. NUMERICAL EXPERIMENTS

In this section several registration tasks, both on synthetic datasets and real medical images, are presented in order to analytically compare MIND and G-MIND.

*A. Distorting transformation and compensating registration on synthetic dataset*

Prior the experiment, a synthetic dataset was produced, including 10 images. The dataset was generated by taking a single underlying image assembled from various segments. Each of the assembled segments was deformed by several types of blurring filters and additive noise, applied on each image segment separately.

At first, 15 pairs of images from this synthetic dataset were randomly selected. For each pair, one image was arbitrarily defined as the reference image and the other as the test image. Since each pair of images describes the same underlying object, both images are completely spatially compatible.

In the second step, an affine transformation $T_{distortion}$ was applied on the test image, resulting in a displacement $\mathbf{U}_{distortion}$. After the affine transformation was applied, the test image is no longer fit the reference image, as demonstrated in Figure 2. Every pair from the synthetic dataset was registered twice, by MIND and by G-MIND, each yielding its own registration displacement $\mathbf{U}_{registration}$.

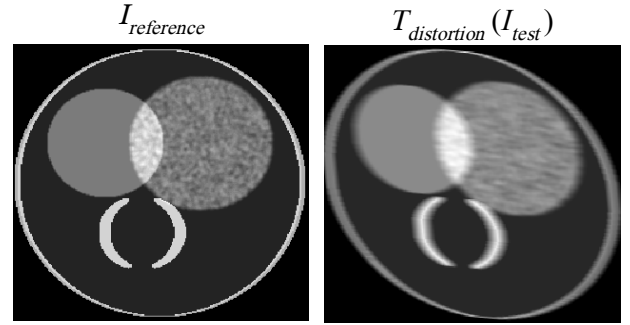

Fig. 2. An example of a pair of images from the synthetic dataset presented in Section IV. The left image was arbitrarily chosen as a reference, while the right is used as a test image and was distorted by $T_{distortion}$. This pair is used as the source images in the experiment.

We define the registration error as:

$$error(\mathbf{X}) = \left\|\mathbf{U}_{registration}(\mathbf{X}) + \mathbf{U}_{distortion}(\mathbf{X})\right\|_2. \quad (9)$$

A perfect registration should completely compensate for the distortion and thus the error, as defined in Equation 9, is a good measure for estimating the registration accuracy.

For all the 15 synthetic pairs which were registered, G-MIND achieved a significant lower error values on the edges. Figure 3 illustrates a typical result for this experiment. Therefore, in addition to the improvement G-MIND shows on real medical images (see Section III and Figure 1), it can be concluded that G-MIND outperforms MIND accuracy for edges.

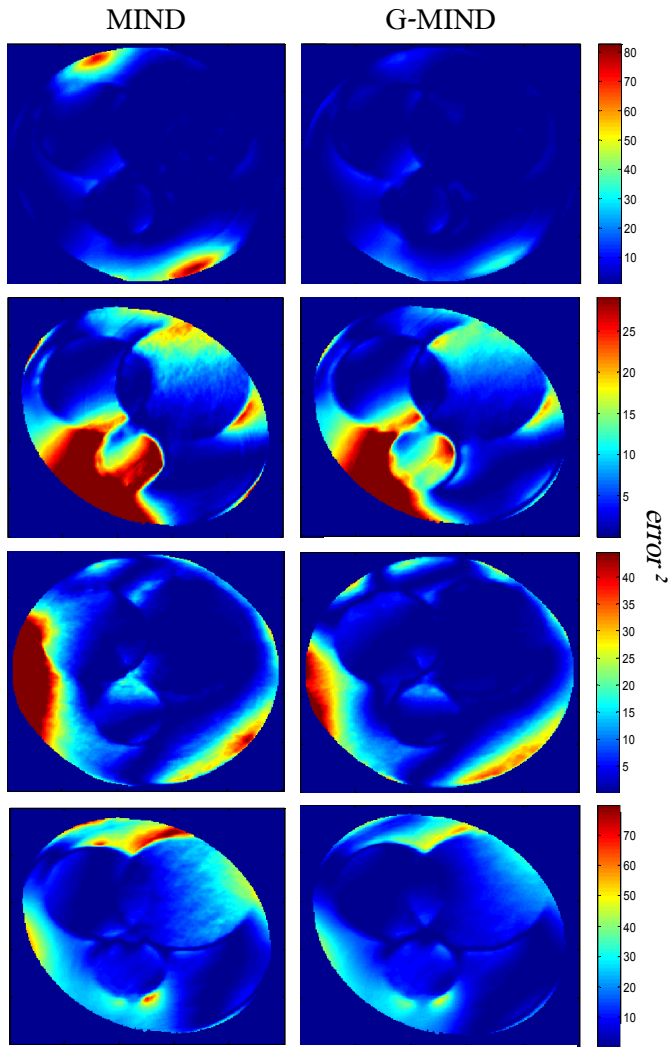

Fig. 3. This figure presents the error of the displacement field **U**(x,y) for MIND and G-MIND registrations. Each line presents a different synthetic pair. The two upper lines present the displacement's error in x-direction, while the two bottom lines present the y-direction. Low values present small registration error. G-MIND appears to be more accurate on the object edges, which can be located according to Figure 2.

### B. Landmark localization on inhale and exhale CT scans

We performed a complementary experiment of regional landmark localization in order to examine G-MIND accuracy on other image textures and features. Five pairs of thorax and upper abdomen breathing cycle CT scans were provided by the DIR-Lab at the University of Texas [3]. The challenge of this single-modal registration tasks stems from the changes during breathing cycle: contrast changes due to the change of gas density, motion of the lung lobes and ribs and large deformations of small features [4].

For each pair of breathing cycle CT scans, 300 corresponding anatomical landmarks were delineated by thoracic imaging expert with a mean selection error of 1 mm. The corresponding landmarks are located at a variety of image textures, excluding edges. An example of a corresponding landmarks pair is showed in Figure 4. Every pair was registered twice, by MIND and by G-MIND.

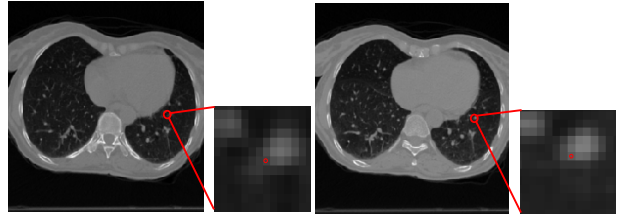

Fig. 4. An example of corresponding landmarks from pair number 2 of 4DCT DIR-LAB dataset.

A common way to estimate the registration accuracy is by calculating the distance between corresponding landmarks after registrations, known as the Target Registration Error (TRE) [11]. Given two landmarks, $\mathbf{X}_1=(x_1, y_1)^T$, $\mathbf{X}_2=(x_2, y_2)^T$, the TRE is calculated as:

$$TRE = \sqrt{(x_1 + u_x(x_1) - x_2)^2 + (y_1 + u_y(y_1) - y_2)^2}, \quad (10)$$

where $\mathbf{U}=(u_x, u_y)^T$ is the registration displacement from one landmark to another.

TABLE I.  LANDMARK LOCALIZATION RESULTS

| Registration method | TRE [pixels]<br>mean ± std | TRE [mm]<br>mean ± std |
|---|---|---|
| MIND | 1.01 ± 2.76 | 1.48 ± 4.04 |
| G-MIND | 1.19 ± 3.36 | 1.98 ± 4.71 |

Table 1 shows the results of both methods for the five pairs of CT scans (1500 landmarks in total). Even though MIND achieved a slightly lower mean TRE (1.48 mm comparing to 1.98mm), the difference is small compared to the landmarks selection mean error. Additionally, in pixel terms, the mean difference between the TREs is lower than a fifth of a pixel. That is to say, the difference is indistinguishable compared to the image resolution.

This experiment confirms that along with the major improvement in edge preserving registration, G-MIND does not degrade the original MIND registration performance for other textures and features.

## V. DISCUSSION AND CONCLUSION

In this work, we have presented a new multi-modal registration method, incorporating MIND approach and gradient intensity. The proposed method shows an accurate registration for several medical imaging challenges, particularly addressing edge preserving. As aforementioned, reliable edge registration may be significantly meaningful for a variety of clinical diagnosis procedures. The G-MIND capabilities were demonstrated by extensive numerical experiments. The results plainly emphasize the advantages of our method in edge registration.

While our work was focused on CT and MRI scans, we strongly believe that G-MIND can be applied to other modalities and applications. This could be a subject for a future research.

Further improvement might be possible by integrating MIND and G-MIND into a single framework by adaptively combining their displacements in order to benefit from the strengths of both methods.

ACKNOWLEDGMENT

The authors thank Dr. Arie Nakhmani, the Control and Robotics Laboratory (CRML) and the Signal and Image Processing Laboratory (SIPL) at the Technion, for supporting this work and SIPL for funding it.